\documentclass[letterpaper, 10 pt, conference]{ieeeconf}
\IEEEoverridecommandlockouts
\usepackage{moreverb,url}
\usepackage[colorlinks,bookmarksopen,bookmarksnumbered,citecolor=red,urlcolor=red]{hyperref}
\usepackage{siunitx}
\usepackage{url}
\usepackage{svg}
\usepackage{multicol}
\usepackage{cprotect}
\usepackage{cellspace} 
\usepackage{soul}
\usepackage{color}
\usepackage{times}
\usepackage{graphicx}
\usepackage{lipsum}
\usepackage{setspace}
\usepackage{epsfig}
\usepackage{amsmath}
\usepackage{bm}
\usepackage{amssymb}
\usepackage{tabularx}
\usepackage{pifont}
\usepackage{mathtools}
\usepackage{color,soul}
\usepackage{paralist}
\usepackage{tikz}
\usepackage{float}
\usepackage{colortbl}
\usepackage{pdfpages}
\usepackage{makecell}
\usepackage{array}
\usepackage{subcaption}
\usepackage{enumerate}
\usepackage{pgfgantt}
\usepackage{amsfonts}
\usepackage{textcomp}
\usepackage{multirow,diagbox}
\usepackage{threeparttable}
\usepackage{booktabs}
\usepackage{bm}
\let\labelindent\relax
\usepackage[inline]{enumitem}
\usepackage[ruled,vlined]{algorithm2e}

\newcolumntype{L}[1]{>{\raggedright\let\newline\\\arraybackslash\hspace{0pt}}m{#1}}
\newcolumntype{C}[1]{>{\centering\let\newline\\\arraybackslash\hspace{0pt}}m{#1}}
\newcolumntype{R}[1]{>{\raggedleft\let\newline\\\arraybackslash\hspace{0pt}}m{#1}}

\newcommand{\SE}{\mathrm{SE}(3)}

\newcommand{\campose}[2]{\prescript{#1}{}{\mathbf{X}}_{#2}}
\newcommand{\objpose}[2]{\prescript{#1}{}{\mathbf{L}}_{#2}}
\newcommand{\worldf}{W}

\newcommand{\objmotion}[3]{\prescript{#1}{#2}{\mathbf{H}}_{#3}}
\newcommand{\objf}{L}
\newcommand{\camf}{X}

\newcommand{\mpoint}[2]{\prescript{#1}{}{\mathbf{m}}_{#2}}

\newcommand{\calcL}[2]{{\objmotion{\worldf}{#1}{#2} \: \objpose{\worldf}{#1} }{}}
\newcommand{\zthreed}{\mathbf{z}_{\text{3D}}}
\newcommand{\factor}[2]{\lVert {#1} \rVert^2_{\Sigma_{{#2}}}}

\usepackage{xcolor}  

\newcommand{\etal}{\textit{et al}.~}

\def\secref#1{Section~\ref{#1}}
\def\figref#1{Fig.~\ref{#1}}
\def\tabref#1{Table~\ref{#1}}
\def\eqref#1{(\ref{#1})}

\title{\LARGE \bf
Online Dynamic SLAM with Incremental Smoothing and Mapping
}

\author{Jesse Morris, Yiduo~Wang and~Viorela~Ila
\thanks{Jesse Morris, Yiduo Wang and Viorela Ila are with the Australian Centre For Robotics (ACFR), University of Sydney, 2006 Sydney, Australia.
{\tt \{jesse.morris,yiduo.wang,viorela.ila\}@sydney.edu.au}}
}

\begin{document}

\maketitle
\thispagestyle{empty}
\pagestyle{empty}

\begin{abstract}

Dynamic SLAM methods jointly estimate for the static and dynamic scene components, however existing approaches, while accurate, are computationally expensive and unsuitable for online applications. 
In this work, we present the first application of incremental optimisation techniques to Dynamic SLAM. 
We introduce a novel factor-graph formulation and system architecture designed to take advantage of existing incremental optimisation methods and support online estimation. 
On multiple datasets, we demonstrate that our method achieves equal to or better than state-of-the-art in camera pose and object motion accuracy.
We further analyse the structural properties of our approach to demonstrate its scalability and provide insight regarding the challenges of solving Dynamic SLAM incrementally.
Finally, we show that our formulation results in problem structure well-suited to incremental solvers, while our system architecture further enhances performance, achieving a $\mathbf{5\times}$ speed-up over existing methods.
\end{abstract}



\section{Introduction}

To build a complete mental map of complex dynamic environments, robotic systems must comprehensively understand \textit{where} an object is and \textit{how} it is moving.
Dynamic objects are inherently unpredictable and therefore \textit{online} estimation of their motion is fundamental to any safety critical operation.
While recent progress in Dynamic Simultaneous Localisation and Mapping (Dynamic SLAM)~\cite{judd2024ijrr_mvo, morris2025dynosam} facilitates highly accurate object motion and pose estimation alongside ego-motion, current methods are computationally complex and limited to intensive batch optimisation, making them unsuited for real-world applications in highly dynamic environments. 
Rather than relying on batch methods, online estimation can be achieved using incremental optimisation algorithms such as iSAM~\cite{kaess2008isam} or iSAM2~\cite{kaess2012isam2}, which exploit the sparsity of the underlying optimisation problem to enable efficient incremental inference~\cite{dellaert2017factor}.
The Dynamic SLAM problem is inherently more complex than the static case, with additional  connections among object points, their motions and the observing camera poses creating a denser optimisation problem that is challenging to solve efficiently.
Thus, formulating a factor-graph for Dynamic SLAM that enhances the problem's underlying sparsity, while maintaining accuracy, is vital to achieve accurate online performance.

Motivated by these needs, we propose a novel \textit{Hybrid} formulation for Dynamic SLAM, 
which merges the benefits of two key representations identified in the literature the \textit{object-centric} and \textit{world-centric} representations~\cite{morris2025dynosam, bescos2021ral, morris2024icra},
Our approach retains the accuracy of world-centric methods, where the rigid-body motion of each object is explicitly modelled, while using the object-centric representation to minimise the number of state variables and reduce overall graph connectivity~\cite{morris2024icra}. 
In additional to estimating for the camera pose, static structure, per-object motion and dynamic map, our formulation naturally accumulates the object structure over time and allows the object pose and velocity at each frame to be directly recovered. 

\begin{figure}[t]
	\centering
	\includegraphics[trim={0.0cm 14.9cm 0.0cm 0cm},clip,width=1.0\columnwidth]{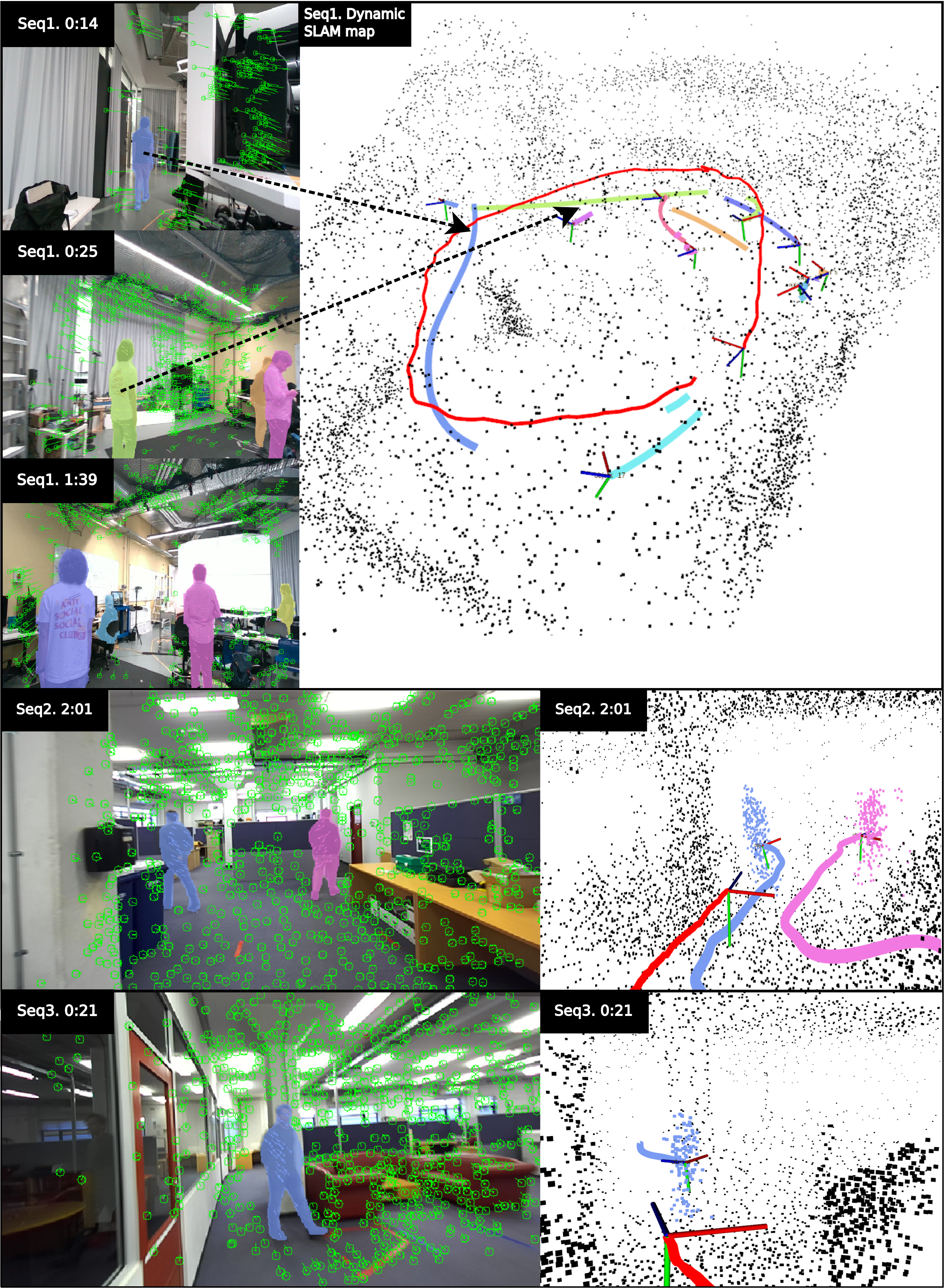}
	\caption{\small{Output of our proposed Dynamic SLAM system recorded in indoor environments with multiple moving objects. Our proposed method employs a novel Hybrid representation for dynamic objects and motions, allowing the camera pose (red trajectory) and static scene (black point cloud) as well as the motion, structure and pose of dynamic objects (uniquely coloured) to be estimated jointly and incrementally using iSAM2~\cite{kaess2012isam2}. We show the 2D tracking on the camera images alongside the 3D estimations.}}
    \label{fig:ecmr_frontpage}
    \vspace{-7mm}
\end{figure}
The proposed Hybrid approach significantly enhances the sparsity of the underlying optimisation problem. However, as the number of visible objects increases, the connectivity of the problem grows accordingly, which can limit overall performance.
To address this, we additionally propose a system architecture that explicitly reduces connectivity between camera and object variables; this enhances the formulation's existing sparsity to better support online incremental inference.
We demonstrate that our method achieves at least a $5\times$ speed-up with minimal to no accuracy loss when compared to a state-of-the-art baseline on real and simulated datasets, and qualitative results on real-world sequences with multiple dynamic (non-rigid) objects demonstrate robustness under realistic conditions, as shown in~\figref{fig:ecmr_frontpage}.

Thus our contributions are:
\begin{itemize}
    \item A novel \textit{Hybrid} formulation for Dynamic SLAM that combines the benefits of the existing object and world-centric approaches.
    \item A novel architecture for Dynamic SLAM that facilitates online and incremental estimation. To the best of our knowledge, this is the first work to apply incremental optimisation methods to the Dynamic SLAM problem. 
    \item We present a detailed comparison with other state-of-the-art Dynamic SLAM algorithms, showing our approach is scalable and more efficient for incremental solvers, thereby better supporting online estimation. 
\end{itemize}

\section{Related Works}


\subsection{Graph-based Dynamic SLAM}
Dynamic SLAM systems seek to incorporate measurements of dynamic objects into the estimation problem. This contrasts with methods that try and improve the quality of camera pose estimation by removing dynamic object; however any dynamic object information is lost~\cite{Bescos2018ral, Campos2021tro, song2024dynavins++}.
Recent graph-based optimisation methods track features on each object and apply a rigid-body motion model to jointly estimate for the object pose/motion and camera poses~\cite{judd2024ijrr_mvo, morris2025dynosam, zhang2020vdoslam, Liu2024ral_dynameshslam, Tian2024its_dynaquadric}. 
Morris~\etal\cite{morris2024icra} categorise Dynamic SLAM methods as object or world-centric and demonstrate how each formulation can significantly affect optimisation convergence and estimation accuracy. 
Therefore, reviewing these formulations is essential to better understand ways to solve Dynamic SLAM incrementally.

\textbf{Object-Centric Representations}~\cite{bescos2021ral, Huang2019iccv, huang2020cvpr, Ballester2021icra} represent dynamic points in the object’s body frame. 
Under a rigid-body assumption each point can be treated as static, and therefore modelled as a single state variable.
While DynaSLAM II~\cite{bescos2021ral} explicitly describe the benefits of this representation on reducing the number of state variables, they do not directly compare or analyse its computational benefits.

Furthermore, many object-centric methods report limited object motion/pose accuracy.
Morris~\etal\cite{morris2024icra} attribute this to the lack of explicit rigid-body kinematics, with most approaches relying on a constant velocity model to constrain the optimisation and the object-centric representation to (implicitly) enforce rigidity.
MVO~\cite{judd2024ijrr_mvo} achieves the most accurate results 
using explicitly point-pair distance constraints to enforcing rigidity and a constant velocity prior to ensure physically plausible trajectories.

\textbf{World-Centric Representations}~\cite{morris2025dynosam, zhang2020vdoslam,Henein20icra} represent object motions and points with respect to a common global frame~\cite{Chirikjian17idetc}.
This motion parametrisation explicitly enforces rigid-body kinematics without the need for additional constraints.
More details on this parametrisation can be found in~\cite{morris2025dynosam, Chirikjian17idetc}.
Recent work DynoSAM~\cite{morris2025dynosam} uses this approach to estimate for the per-frame motion of each object and achieves highly accurate results. 
However, as both motions and points must be explicitly modelled at every time-step, the resulting factor graph is densely connected with many variables~\cite{morris2024icra} leading to computationally heavy batch optimisation. 

\subsection{Incremental Dynamic SLAM}
\label{sec:incremental_slam}
Many state-of-the-art static SLAM systems~\cite{Rosinol20icra, Forster2017tro_svo, Shan2020iros_liosam} employ iSAM2~\cite{kaess2012isam2} to facilitate online localisation and mapping, thereby better supporting robotic exploration. 
Applying incremental optimisation techniques enables efficient updates to the estimation, minimising per-frame computation time while maintaining optimality. 
However, to the best of the authors' knowledge, no work has applied incremental optimisation to the Dynamic SLAM problem.

\section{Preliminaries}

\subsection{Notation}
\label{sec:notation}



Consider a dynamic scene comprised of camera poses $\mathcal{X}$ and object poses $\mathcal{L}$:
\begin{equation*}
   \mathcal{X} = \{ \campose{\worldf}{k} \in \SE \}_{{k \in \mathcal{K}}}, \quad \mathcal{L} = \{ \objpose{\worldf}{k}^j  \in \SE \}{\substack{{j \in \mathcal{J}_k} \\ {k \in \mathcal{K}_j}}}
\end{equation*}
where $\{\worldf\}$ is the world frame, $\mathcal{K}$ is the set of all time-steps and $\mathcal{J}$ is the set of all object indices. 
$\mathcal{J}_k \subseteq \mathcal{J}$ are object indices observed at $k$, and $\mathcal{K}_j \subseteq \mathcal{K}$ is the subset of time-steps over which object $j$ is observed.
Each pose $\campose{\worldf}{k}$ and $\objpose{\worldf}{k}$ are associated with body-fixed reference frames $\{ \camf_k\} $ and $\{\objf_k\}$.  
At each $k$, the camera provides surface measurements, e.g. from RGBD camera, $\mathcal{Z}_{k} = \{ \mathcal{S}_{k}, \mathcal{D}_{k} \}$.
$\mathcal{S}_{k}$ and $ \mathcal{D}_{k} = \{ \mathcal{D}^j_{k} \}^{j \in \mathcal{J}_k}$ are the sets of all static and dynamic point measurements respectively.
The homogenous coordinate of a point $\tilde{\mathbf{m}}^i\in\mathbb{R}^3$ is $\mpoint{}{}^{i} = \left[\tilde{\mathbf{m}}^i, 1\right]^\top$, where $i$ indicates correspondences across frames.
Any point measurement in the sensor frame $\{\camf\}$ is denoted as $\zthreed \in \mathbb{R}^3$ such that:
\begin{equation*}
     \left[\zthreed, 1\right]^\top = \mpoint{\camf_k}{k}^{i} = \campose{\worldf}{k}^{-1} \: \mpoint{\worldf}{k}^{i}\text{.}
\end{equation*}
To minimise notation clutter, we liberally omit indices $i$ and $j$ when there is no ambiguity.

Finally, the motion of object $j$ between time-steps $A$ and $B$ represented in the \textit{world} frame is $\objmotion{\worldf}{A}{B} \in \SE$.
While a pose specifies a body’s position and orientation in a given reference frame, 
a motion maps one pose to another~\cite{Chirikjian17idetc}:
\begin{equation}
    \objpose{\worldf}{B} = \objmotion{\worldf}{A}{B} \: \objpose{\worldf}{A}
\label{equ:object_motion_from_pose_def}
\end{equation}
We highlight that, in contrast to the body-frame motion $\objmotion{\objf_{A}}{A}{B}$ which may roughly be thought of as a velocity, $\objmotion{\worldf}{A}{B}$ lacks an equivalent physical interpretation and should be thought of simply as a transformation that moves a rigid body (and all its points) through space.

Using this motion, any and all tracked points $i$ on a single rigid body $j$ at two different time-steps can be related to each other using a single transform, represented in $\{\worldf\}$:
\begin{equation}
    \mpoint{\worldf}{B} = \objmotion{\worldf}{A}{B} \: \mpoint{\worldf}{A}\text{,}
    \label{equ:general_point_motion}
\end{equation}
This parametrisation of motion is key to existing world-centric methods~\cite{morris2025dynosam, zhang2020vdoslam, Henein20icra} and similarly underpins our approach.
However, we highlight that prior works exclusively consider the consecutive (per-frame) motion $\objmotion{\worldf}{k-1}{k}$ of each object. 
In this work we expand the utility of this representation to \textit{non-consecutive} frames, 
and demonstrate that this allows for a more efficient formulation of the Dynamic SLAM problem.

\subsection{Frontend}
\label{sec:frontend}
This work focuses on the backend formulation for a Dynamic SLAM system, as well as the accuracy and timing capabilities of the proposed formulation and architecture.
We use the frontend implemented in DynoSAM~\cite{morris2025dynosam} to provide frame-to-frame measurements $\mathcal{Z}_k$.


\subsection{Incremental Inference \& iSAM2 in Dynamic SLAM}
\label{sec:isam2_summary}

\begin{figure}[b]
    \vspace{-4mm}
	\centering
	\includegraphics[trim={0.0cm 0.0cm 0.0cm 0cm},clip,width=1.0\columnwidth]{figs/bayes_tree_compressed.pdf}
	\caption{\small{A possible Bayes Tree generated using the world-centric formulation proposed in~\cite{morris2024icra} from a scene containing $3$ objects observed over $4$ frames. Notice that following the iSAM2 algorithm creates large cliques which contains variables from all objects (highlighted in red), thus leading to inefficient computation.}}
    \label{fig:bayes_tree_example}
\end{figure}

ISAM2~\cite{kaess2012isam2} is a non-linear, graph-based optimisation algorithm that facilitates efficient online updates by using the Bayes Tree~\cite{kaess2010bayes} to exploit the sparsity structure of the underlying system.
A Bayes Tree $\mathcal{B}$ is built from a chordal Bayes Net via variable elimination on a factor graph $\mathcal{F}$, with each node representing a clique of conditionally dependent variables. 
Inference is performed via variable elimination of each clique in bottom-up order; this is equivalent to forming the Cholesky factor $\mathbf{R}$ for the linearised $\mathcal{F}$~\cite{dellaert2017factor}.
The sparsity structure of $\mathbf{R}$ is directly captured in the topology of $\mathcal{B}$ where large cliques indicate that $\mathbf{R}$ is dense and solving the system is therefore highly inefficient.
Consequently, the memory and computation performance of inference depends directly on the sparsity structure of $\mathbf{R}$, 
and designing $\mathcal{F}$ to preserve or enhance this sparsity is therefore critical for efficient, incremental estimation.

Instead of performing inference over the entire Bayes Tree (equivalent to batch each step), iSAM2 achieves fast incremental updates by only recomputing parts of $\mathcal{B}$ affected by new measurements. 
By ordering recently affected variables near the root, computation is confined to the top of the tree. 
Static SLAM typically contains small cliques~\cite{dellaert2017factor}, and therefore a sparse underlying representation, that can be solved and updated efficiently. 
In comparison, \figref{fig:bayes_tree_example} presents the Bayes Tree created by naively applying iSAM2 to a Dynamic SLAM formulation~\cite{morris2024icra} in a toy scenario where multiple objects are continuously observed. 
Observe that there are large cliques that contain variables from many or all objects present in the scene, as highlighted in red in~\figref{fig:bayes_tree_example}, making inference inefficient and computation scale with the number of dynamic objects' variables. 
Similar challenges arise in multi-robot SLAM~\cite{zhang2021mr}.
To mitigate this, we propose a formulation that preserves sparsity during incremental updates in the following section.


\section{Method}
This section presents our novel formulation for Dynamic SLAM to facilitate efficient iSAM2 optimisation.
Our \textit{Hybrid} approach combines the advantages of object and world-centric methods and we discuss the additional benefits of our approach compared to existing formulations.

\subsection{Hybrid Representation For Dynamic SLAM}

\begin{figure}[t]
	\centering
	\includegraphics[trim={0.0cm 0.0cm 0.0cm 0cm},clip,width=0.9\columnwidth]{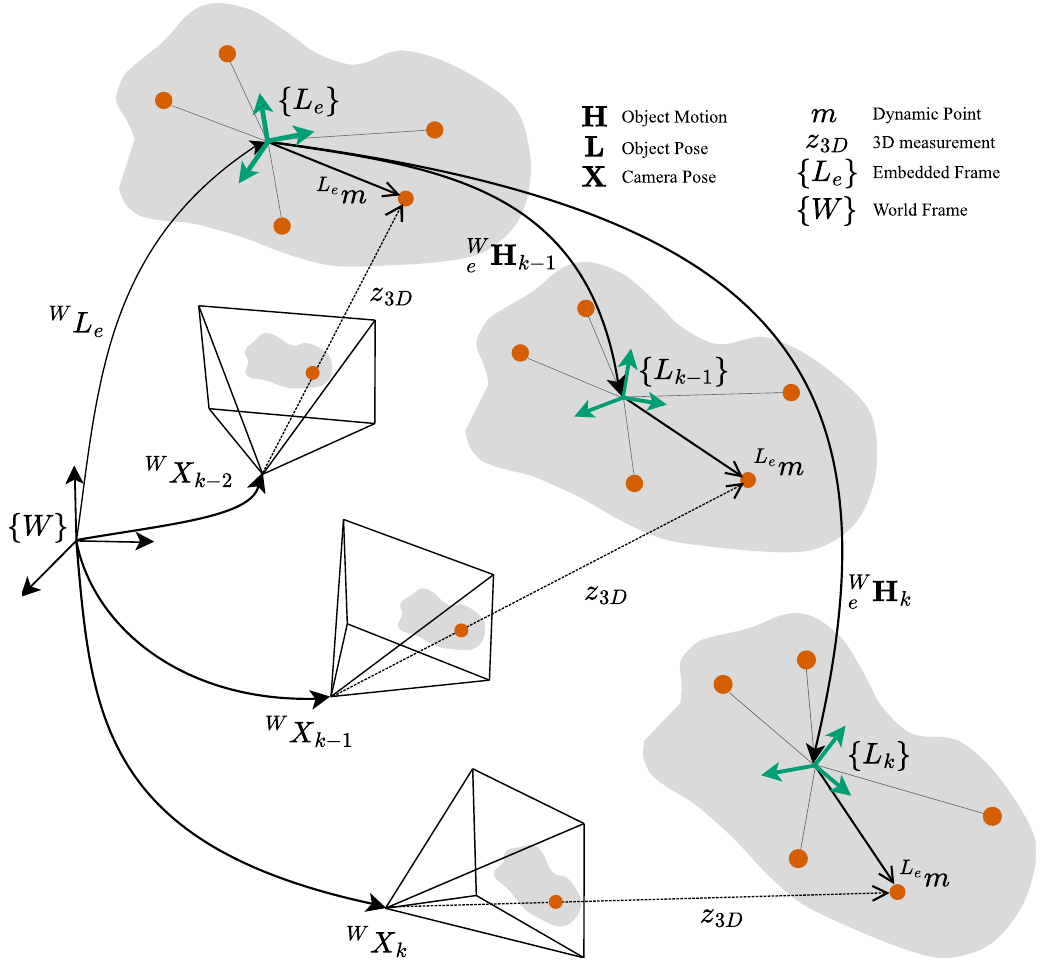}
	\caption{\small{Our \textit{Hybrid} Dynamic SLAM representation combines the benefits of an object-centric point representation and a world-centric representation for motions. Here we show a object with $5$ points seen at three consecutive frames. The first frame, $k-2$, is used to define the objects embedded frame $\{\objf_e\}$ which is moved through \textit{time and space} by the estimation object motion $\objmotion{\worldf}{e}{k}$. For simplicity, we notate only one point $\mpoint{\objf_{e}}{}$ on the object, as well as its associated measurement $\zthreed$.}}
    \label{fig:ecmr_2025_notation}
    \vspace{-7mm}
\end{figure}

Our approach considers the motion of each object as a transformation of the body-frame $\{\objf_k\}$ relative to some embedded frame $\{\objf_e\}$. 
This frame is explicitly defined for all objects $j \in \mathcal{J}$ when they are first observed, i.e. ${k=e}$, and is \textit{fixed} with respect to $\{ \worldf\}$.
As illustrated in~\figref{fig:ecmr_2025_notation}, $\{\objf_e\}$ serves as the common frame of reference \textit{for all associated points and future motions.}
Object points are expressed directly in the body frame as static variables $\mpoint{{\objf_e}}{}$.
The object motion is defined as $\objmotion{\worldf}{e}{k}$, which acts as a relative world-centric transformation, transporting the embedded frame $\{\objf_e\}$, along with all attached points, from $e$ to $k$ while explicitly enforcing rigid-body constraints.

The key novelty of our approach is to anchor each rigid-body motion on the explicit object frame $\{\objf_e\}$.
Compared to world-centric methods that lack an explicit object frame, 
using $\{\objf_e\}$ as a common reference for all object points allows the object map to naturally grow overtime, leading to a more complete representation as new object fragments become visible. 
For each point $\mpoint{{\objf_e}}{}$ in the object map, its global position is found by propagating a point in $\{\objf_e\}$ \textit{through time and space} to $\{\worldf\}$ at time-step $k$ following~\eqref{equ:general_point_motion}:
\begin{equation}
    \mpoint{\worldf}{k} = \objmotion{\worldf}{e}{k} \: \objpose{\worldf}{e} \: \mpoint{{\objf_e}}{}\text{,}
\label{equ:local_point_to_world}
\end{equation}
where the pose $\objpose{\worldf}{e}$ is associated with the objects embedded frame.
Applying~\eqref{equ:local_point_to_world} to all object points recovers the most up-to-date map for any $k \in \mathcal{K}_j$.

Explicitly defining $\{\objf_e\}$ further allows the object pose to be recovered directly for any $k \in \mathcal{K}_j$ given the motion:
\begin{equation}
    \objpose{\worldf}{k} = \calcL{e}{k}\text{.}
\label{equ:object_pose_calc}
\end{equation}
This avoids explicitly estimating the object pose, as required in object-centric methods~\cite{bescos2021ral, gonzalez2022twistslam}, or recursively propagating per-frame motions for pose calculation as in world-centric approaches~\cite{morris2025dynosam}.
A key implication of this approach is that all object poses are anchored to $\{\worldf\}$ via $\{\objf_e\}$, as shown in~\figref{fig:ecmr_2025_notation}, making the entire trajectory dependent on the choice of ${\objf_e}$.
Though this may seem limiting, prior work~\cite{morris2025dynosam, morris2024icra, Chirikjian17idetc} shows that world-centric motion representations allows $\{\objf_e\}$ to be defined \textit{arbitrarily} without affecting estimation.
Such flexibility supports various initialisation strategies, including learned priors or alignment with known models.
In our implementation, we define $\{\objf_e\}$ at the centroid of the first measurements with identity rotation.

After recovering object poses using~\eqref{equ:object_pose_calc}, the body-frame motion is computed as:
\begin{equation}
    \objmotion{\objf_{k-1}}{k-1}{k} = \objpose{\worldf}{k-1}^{-1} \: \objpose{\worldf}{k}\text{,}
    \label{equ:local_object_motion_calc}
\end{equation}
from which the linear and angular velocity can be obtained.
\begin{figure}[t]
	\centering
	\includegraphics[clip,width=0.84\columnwidth]{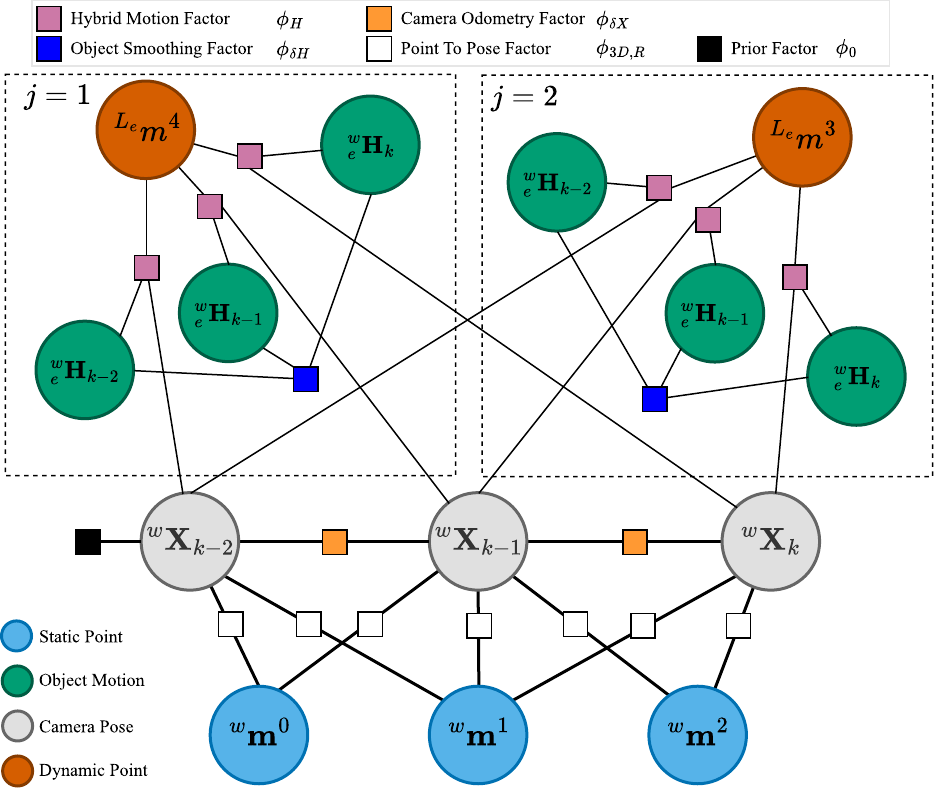}
	\caption{\small{Full Hybrid Dynamic SLAM factor-graph showing static points (blue) and camera poses (grey) at three consecutive frames. Two dynamic objects are shown, each with one point (orange), observed in all frames and connected by the proposed hybrid motion factors ($\phi_H$). The ternary object smoothing factors ($\phi_{\delta H}$) constrains the change in object motion (green). } }
    \label{fig:hybrid_fg}
    \vspace{-6mm}
\end{figure}

\subsection{Factor Graph Construction}
\label{sec:full_hybrid}
Using the Hybrid formulation, 
we pose our Dynamic SLAM estimation as a maximum-a-posteriori (MAP) problem using factors in the form of:
\begin{equation}
    \phi(\cdot) \propto \text{exp} \Bigl\{ -\frac{1}{2} \factor{\mathbf{r}}{} \Bigr\}
\label{equ:factor_form}
\end{equation}
where the residual function $\mathbf{r}$ has a corresponding covariance matrix $\Sigma$.
This section outlines our novel residual functions associated with dynamic measurements and states.
The complete factor graph is shown in~\figref{fig:hybrid_fg} where the static SLAM component uses the factors and method outlined in \cite{morris2025dynosam}. 

To enforce the rigid-body motion of each dynamic point we introduce the \textit{hybrid motion factor}:
\begin{equation}
    \mathbf{r}_{H} = \zthreed - \campose{\worldf}{k}^{-1} \: \objmotion{\worldf}{e}{k} \: \objpose{\worldf}{e} \: \mpoint{{\objf_e}}{}\text{,}
\label{equ:object_centric_point_factor}
\end{equation}
where $\zthreed$ is a measurement of the point $\mpoint{{\objf_e}}{}$, $\objmotion{\worldf}{e}{k}$ is the rigid-body motion of object $j$ and $\campose{\worldf}{k}$ is the observing camera pose. Importantly, $\objpose{\worldf}{e} $ is fixed for each object.
New points seen at $k$ are initialised by projecting the measurement into the embedded object frame:
\begin{equation}
    \mpoint{\objf_e}{} = \objpose{\worldf}{e}^{-1} \: \objmotion{\worldf}{k}{e} \: \campose{\worldf}{k} \: \zthreed
\label{equ:oject_centric_point_projection}
\end{equation}
In the case that $k > e$, $\objmotion{\worldf}{k}{e}$ moves the point backwards from $k$ to $e$. 
Note that this `reversing' motion is not equivalent to inverting $\objmotion{\worldf}{e}{k}$~\cite{Chirikjian17idetc}, and is instead calculated as:
\begin{equation}
     \objmotion{\worldf}{k}{e} = \objpose{\worldf}{k} \: \big(\objpose{\worldf}{e}^{-1} \: \objmotion{\worldf}{e}{k} \:  \objpose{\worldf}{e} \big)^{-1} \: \objpose{\worldf}{k}^{-1}
\label{equ:inverse_motion_calc}
\end{equation}
Furthermore, we incentivise the estimation of physically plausible motions via a constant motion model:
\begin{equation}
        \mathbf{I} =\objmotion{\objf_{k-2}}{k-2}{k-1}^{-1} \: \objmotion{\objf_{k-1}}{k-1}{k}\text{,}
\label{equ:constant_motion_local_frame}
\end{equation}
where~\eqref{equ:local_object_motion_calc} is used to calculate the relative motions $\objmotion{\objf_{k-2}}{k-2}{k-1}$ and $\objmotion{\objf_{k-1}}{k-1}{k}$.
Rewriting~\eqref{equ:constant_motion_local_frame} using the estimated object motions forms a ternary \textit{object smoothing factor}:
\begin{equation}
\begin{aligned}
    \mathbf{r}_{\delta H} = \log  \Bigg[& \left( \left( \calcL{e}{k-2} \right) ^{-1} \:  \left( \calcL{e}{k-1} \right) \right)^{-1} \\
    &\left(\left( \calcL{e}{k-1} \right) ^{-1} \:  \left( \calcL{e}{k} \right) \right)  \Bigg] ^\vee    
\label{eq:smoothing_factor}
\end{aligned}
\end{equation}

This constraint is analogous to a constant velocity prior when the time between frames is constant.
In contrast to world-centric formulations that define the residual in $\{\worldf\}$, our residual in~\eqref{eq:smoothing_factor} is defined in the object’s body-frame.
As discussed in~\cite{morris2025dynosam}, computing this residual in the world-frame causes it to scale with the object’s distance from the origin, which can distort the estimation by overemphasising errors for faraway objects.
Using the body-frame avoids this issue and also allows class-specific motion models, such as those utilised in~\cite{gonzalez2022twistslam, judd2024ijrr_mvo}, to be more naturally expressed.

\textbf{Remark:} Our formulation is explicitly designed to preserve and enhance sparsity when solving incrementally.
As shown in~\figref{fig:hybrid_fg}, each dynamic point becomes a leaf node in the factor graph, and therefore may be eliminated efficiently without introducing large cliques. 
This makes the Hybrid formulation well-suited for incremental solvers; such improvement is validated by the Bayes Tree analysis in~\secref{sec:bayes_tree_analysis}.
In contrast, world-centric methods often form long chains of interconnected dynamic points, leading to large cliques.

However, the Hybrid formulation solves a joint estimation problem that, as discussed in~\secref{sec:isam2_summary}, results in computation scaling with the number of object variables.
From the factor-graph perspective (\figref{fig:hybrid_fg}), all objects variables are interconnected through shared camera poses that serve as the sole separators between the static and dynamic scene components.
This coupling introduces additional dependencies among variables and leads to the formation of large cliques containing variables from all recently observed objects.

\section{Parallel Hybrid}
\label{sec:parallel_hybrid}
\begin{figure}[t]
	\centering
	\includegraphics[trim={0.0cm 0.0cm 0.0cm 0cm},clip,width=0.99\columnwidth]{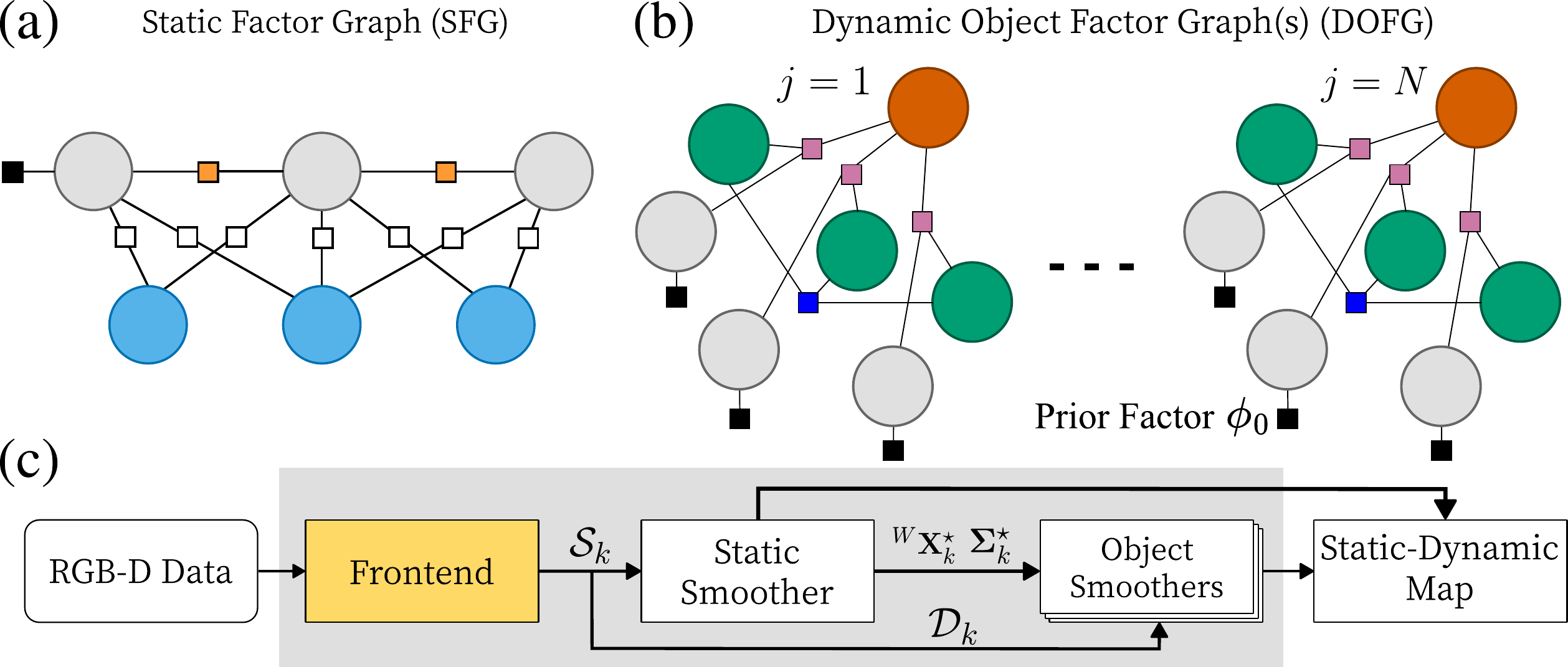}
	\caption{\small{Parallel-Hybrid approach decouples the static \textbf{(a)} and dynamic \textbf{(b)} components from the Hybrid formulation~\figref{fig:hybrid_fg}. \textbf{(c)} shows the system architecture and each smoother solves its corresponding factor-graph. Information is passed from the static factor-graph (SFG) to the dynamic object factor-graphs (DOFG) via the highlighted pose priors $\phi_0$.}}
    \label{fig:parallel_oc_fg}
    \vspace{-6mm}
\end{figure}

To better preserve sparsity during incremental inference and support online inference, we propose a modified variant where the factor graph (\figref{fig:hybrid_fg}) is `cut' along the motion factors, partitioning it a static factor graph (SFG) and $N = \lvert \mathcal{J} \rvert$ dynamic object factor graph's (DOFG), as shown in~\figref{fig:parallel_oc_fg}~(a-b).
Each dynamic object variable is anchored to the world by \textit{conditioning} the estimation on the camera pose, as shown by the inclusion of camera pose variables in each DOFG.
This fully decouples each DOFG from each other and the static scene allowing each system to be solved independently and in parallel; we call this architecture \textit{Parallel-Hybrid}.

To incrementally solve this decoupled structure we propose the system architecture shown in~\figref{fig:parallel_oc_fg}~(c).
We maintain a separate instance of the iSAM2 algorithm for each factor graph (one SFG and $N$ DOFGs).
From new measurements $\mathcal{Z}_k$, the SFG is updated first with factors constructed from $\mathcal{S}_k$, yielding the optimal camera pose $\campose{\worldf}{k}^\star$ and associated marginal covariance $\Sigma_{\campose{}{k}}^\star$.
Each DOFG is then updated with factors constructed from $\{ \campose{\worldf}{k}^\star, \Sigma_{\campose{}{k}}^\star, \mathcal{D}_k^j \}$. 
To condition the camera pose in each DOFG, we add a prior factor $\phi_0$ using the current mean and covariance, as highlighted in~\figref{fig:parallel_oc_fg}~(b).

Including the camera pose as a variable within each DOFG is essential to facilitate incremental updates, as updates to any camera pose will now naturally propagate through the Bayes Tree and affect the estimates of object points and motions.
However, since the SFG and the DOFG's are maintained in separate Bayes Trees, this propagation does not occur automatically. 
Instead, we manually update the mean and covariance of each $\phi_0$ when the corresponding pose $\campose{\worldf}{k}^\star$ undergoes significant relinearisation.
We detect such updates using the fluid relinearisation strategy from iSAM2~\cite{kaess2012isam2}, although this detection method could be refined.
One draw back of the proposed architecture is that information flow is unidirectional and prevents any useful dynamic object information propagating back through the SFG to influence the camera pose estimation.

\section{Experiments}
\label{sec:experiments}

\subsection{Setup}

Due to the limited systems that solve Dynamic SLAM incrementally,
we use the world-centric formulation~\cite{morris2024icra, zhang2020vdoslam, Henein20icra} as implemented in DynoSAM~\cite{morris2025dynosam} as the state-of-the-art Baseline for comparison.
We evaluate on the KITTI tracking~\cite{Geiger13ijrr}, Outdoor Cluster~\cite{Huang2019iccv} and OMD~\cite{Judd19ral} datasets which contain relevant ground truth.
We include the TartanAir Shibuya~\cite{Qiu2022icra_airdos} and VIODE~\cite{minoda2021viode} datasets to further evaluate computation time in highly dynamic environments---these datasets do not provide ground truth object motions/poses.

We evaluate both Hybrid and Baseline formulations using incremental and batch solvers. When solved using iSAM2, the method name is prefixed with `i-'; otherwise, it is solved full-batch using Levenberg–Marquardt. Parallel-Hybrid is always solved incrementally.




The Hybrid and Parallel-Hybrid methods are implemented using the open-source DynoSAM~\cite{morris2025dynosam} framework, 
which uses GTSAM~\cite{gtsam} (version 4.2.0) for factor graph optimisation.
We use a Intel Core i9-9900 CPU with 32GB RAM and 2GB swap memory for all experiments.

\definecolor{lightred}{RGB}{255, 0, 0}
\definecolor{kellygreen}{rgb}{0.3, 0.73, 0.09}
\colorlet{transparentred}{lightred!30} 
\colorlet{transparentgreen}{kellygreen!30} 

\newcommand{\redcell}[1]{\cellcolor{transparentred} $#1$}
\newcommand{\greencell}[1]{\cellcolor{transparentgreen} $#1$}

\begin{table*}[t]
\footnotesize
\centering
\setlength{\tabcolsep}{2.3pt}
\caption{\small{Accuracy of Object Motion and Camera Pose. Baseline and Hybrid methods are solved batch. iHybrid and Parallel-Hybrid are solved incrementally.
The metric error is reported for the Baseline method while other rows report improvement relative to the Baseline in \colorbox{transparentgreen}{green} and relatively worse performance in  \colorbox{transparentred}{red}.$\times$ indicates system failure as discussed in~\secref{sec:bayes_tree_analysis}}. }
\label{tab:object_experiments_reduced}
\begin{tabular}{ccccccccccccccccc}
\toprule
& \multirow{2}{*}{Metric} & \multirow{2}{*}{Method} & \multicolumn{9}{c}{{KITTI}} & \multicolumn{4}{c}{{Outdoor Cluster}} & {OMD} \\
\cmidrule(lr){4-12} \cmidrule(lr){13-16}  \cmidrule(lr){17-17}
& & & 00 & 01 & 02 & 03 & 04 & 05 & 06 & 18 & 20 
& L1 & L2 & S1 & S2 & S4U \\
\midrule

\multirow{8}{*}{\rotatebox[origin=c]{90}{\textbf{Object Error}}}
& \multirow{4}{*}{\rotatebox[origin=c]{0}{$\text{ME}_r$(\si{\degree})}}
& Baseline & ${1.11}$ & ${1.04}$ & ${0.97}$ & ${0.26}$ & ${1.24}$ & ${0.85}$ & ${0.39}$ & ${0.57}$ & ${0.52}$ & ${0.82}$ & ${0.70}$ & ${0.69}$ & $2.36$ & $0.67$ \\
& & Hybrid & \redcell{-0.23} &  \redcell{-0.06}  & \greencell{0.1} & \greencell{0.01} & \greencell{0.15} & \greencell{0.3} & \greencell{0.19} & \greencell{0.27} & \redcell{-0.3} & \redcell{-0.34} & \greencell{0.19} & \greencell{0.16} & \greencell{0.03} & \greencell{0.08} \\
& & iHybrid & \greencell{0.12} & \greencell{0.08} & \redcell{-0.35} & \redcell{-0.03} & \greencell{0.71} & \redcell{-0.10} & \greencell{0.07} & \greencell{0.24} & $\times$ & \redcell{-0.46} & \redcell{-0.4} & \greencell{0.3} & \greencell{0.71} & $\times$ \\
& & Parallel-Hybrid & \redcell{-0.06} & \redcell{-0.41} & \redcell{-1.29} & \redcell{-0.12} & \redcell{-0.23} & \redcell{-0.27} & \redcell{-0.09} & \redcell{-0.34} & \redcell{-0.14} & \redcell{-0.25} & \redcell{-0.18} & \redcell{-0.06} & \greencell{0.3} & \greencell{0.07} \\

\cmidrule(lr){2-17}

& \multirow{4}{*}{\rotatebox[origin=c]{0}{$\text{ME}_t$(\si{\meter})}}
& Baseline & ${0.15}$ & ${0.32}$ & ${0.51}$ & ${0.11}$ & ${0.12}$ & ${0.27}$ & ${0.09}$ & ${0.11}$ & ${0.11}$ & ${0.08}$ & ${0.06}$ & ${0.04}$ & ${0.15}$ & ${0.02}$  \\
& & Hybrid & \redcell{-0.08} & \redcell{-0.05} & \greencell{0.0} & \greencell{0.0} & \greencell{0.0} & \redcell{-0.16} & \redcell{-0.02} & \greencell{0.03} & \greencell{0.0} & \redcell{-0.04} & \redcell{-0.04} & \greencell{0.0} & \greencell{0.11} & \greencell{0.0} \\
& & iHybrid & \redcell{-0.08} & \redcell{-0.08} & \greencell{0.06} & \redcell{-0.01} & \redcell{-0.18} & \redcell{-0.16} & \redcell{-0.11} & \redcell{-0.01} & $\times$ & \redcell{-0.04} & \redcell{-0.05} & \greencell{0.0} & \greencell{0.09} & $\times$ \\
& & Parallel-Hybrid & \greencell{0.01} & \redcell{-0.06} & \greencell{0.0} & \greencell{0.0} & \redcell{-0.04} & \redcell{-0.09} & \greencell{0.01} & \redcell{-0.04} & \greencell{0.0} & \greencell{0.02} & \redcell{-0.01} & \redcell{-0.02} & \redcell{-0.04} & \greencell{0.0} \\

\midrule

\multirow{12}{*}{\rotatebox[origin=c]{90}{\textbf{Camera Error}}}
& \multirow{4}{*}{\rotatebox[origin=c]{0}{ATE(\si{\meter})}}
& Baseline & ${1.54}$ & ${2.10}$ & ${0.74}$ & ${1.64}$ & ${1.28}$ & ${2.01}$ & ${0.41}$ & ${2.30}$ & ${2.30}$ & ${0.62}$ & ${0.52}$ & ${0.09}$ & ${1.08}$ & ${0.10}$ \\
& & Hybrid & \greencell{0.0} & \greencell{0.0} & \greencell{0.0} & \redcell{-0.02} & \redcell{-0.01} & \greencell{0.0} & \greencell{0.0} & \greencell{0.0} & \greencell{0.0} & \greencell{0.0} & \greencell{0.0} & \redcell{-0.06} & \greencell{0.81} & \greencell{0.0} \\
& & iHybrid & \redcell{-0.01} & \greencell{0.0} & \greencell{0.0} & \redcell{-0.19} & \redcell{-0.02} & \greencell{0.01} & \redcell{-0.01} & \redcell{-0.06} & $\times$ & \redcell{-0.02} & \redcell{-0.03} & \redcell{-0.05} & \greencell{0.46} & $\times$ \\
& & Parallel-Hybrid & \redcell{-0.01} & \greencell{0.0} & \greencell{0.0} & \redcell{-0.03} & \redcell{-0.04} & \greencell{0.01} & \redcell{-0.02} & \redcell{-0.1} & \redcell{-0.12} & \greencell{0.16} & \redcell{-0.23} & \redcell{-0.02} & \redcell{-0.3} & \greencell{0.0} \\

\cmidrule(lr){2-17}

& \multirow{4}{*}{\rotatebox[origin=c]{0}{$\text{RPE}_r$(\si{\degree})}}
& Baseline & ${0.05}$ & ${0.03}$ & ${0.02}$ & $0.07$ & ${0.07}$ & ${0.07}$ & ${0.05}$ & ${0.04}$ & ${0.03}$ & ${0.02}$ & ${0.02}$ & ${0.01}$ & ${0.02}$ & ${0.66}$ \\
& & Hybrid & \redcell{-0.01} & \greencell{0.0} & \greencell{0.0} & \greencell{0.1} & \greencell{0.0} & \redcell{-0.01} & \greencell{0.0} & \greencell{0.0} & \greencell{0.0} & \redcell{-0.01} & \redcell{-0.01} & \greencell{0.0} & \greencell{0.0} & \greencell{0.0} \\
& & iHybrid & \greencell{0.0} & \redcell{-0.01} & \greencell{0.0} & \greencell{0.2} & \greencell{0.0} & \redcell{-0.01} & \greencell{0.0} & \greencell{0.0} & $\times$ & \greencell{0.0} & \greencell{0.0} & \greencell{0.0} & \greencell{0.0} & $\times$ \\
& & Parallel-Hybrid & \greencell{0.0} & \greencell{0.0} & \greencell{0.0} & \greencell{0.01} & \greencell{0.02} & \greencell{0.0} & \greencell{0.01} & \greencell{0.0} & \greencell{0.0} & \greencell{0.0} & \greencell{0.01} & \greencell{0.01} & \greencell{0.0} & \greencell{0.0} \\

\cmidrule(lr){2-17}

& \multirow{4}{*}{\rotatebox[origin=c]{0}{$\text{RPE}_t$(\si{\meter})}}
& Baseline & ${0.04}$ & ${0.06}$ & ${0.06}$ & ${0.07}$ & ${0.06}$ & ${0.08}$ & ${0.01}$ & ${0.05}$ & ${0.04}$ & ${0.02}$ & ${0.01}$ & ${0.01}$ & ${0.02}$ & ${0.01}$ \\
& & Hybrid & \greencell{0.0} & \greencell{0.0} & \greencell{0.0} & \greencell{0.0} & \greencell{0.0} & \greencell{0.0} & \greencell{0.0} & \greencell{0.0} & \greencell{0.0} & \greencell{0.0} & \greencell{0.0} & \greencell{0.0} & \greencell{0.01} & \greencell{0.0} \\
& & iHybrid & \greencell{0.00} & \greencell{0.01} & \greencell{0.01} & \redcell{-0.01} & \redcell{-0.01} & \redcell{-0.02} & \greencell{0.00} & \greencell{0.00} & $\times$ & \greencell{0.00} & \greencell{0.00} & \greencell{0.00} & \greencell{0.00} & $\times$ \\
& & Parallel-Hybrid & \greencell{0.00} & \greencell{0.00} & \greencell{0.01} & \greencell{0.00} & \greencell{0.00} & \greencell{0.00} & \greencell{0.00} & \greencell{0.00} & \greencell{0.00} & \greencell{0.01} & \greencell{0.00} & \greencell{0.00} & \redcell{-0.01} & \greencell{0.00} \\

\bottomrule
\end{tabular}
\vspace{-6mm}
\end{table*}

\subsection{Estimation Accuracy}
We evaluate camera pose and object motion estimation for all methods.
To ensure a fair comparison with the Baseline and to isolate the inherent accuracy of our proposed formulation, we report full-batch results for Hybrid.
We also report incremental (iHybrid and Parallel-Hybrid) results to demonstrate intended online performance relative to the batch solution.
Parallel-Hybrid, which decouples the estimation problem, additionally offers a direct comparison to the other methods that solve the problem jointly.
Together, these results offer insights into how the solver type (incremental vs batch) and problem structure (joint vs decoupled) influence estimation accuracy.
No system contains loop-closure.

\tabref{tab:object_experiments_reduced} reports the error of the Baseline for each metric, 
and the relative numerical error differences of all proposed methods compared to the Baseline. 
We report standard camera pose metrics: the RMSE of the absolute trajectory error (ATE) and relative pose error (RPE). 
For objects we report RMSE of motion error (ME)~\cite{morris2025dynosam}, averaged over all objects in the sequence.
ME is a reliable way to report motion error as it is impartial to the object frame definition, which may differ among systems.


All methods achieve accuracy comparable to the Baseline across all metrics.
In RPE, errors differ by no more than \SI{0.02}{\meter} (translation) and \SI{0.01}{\degree} (rotation), while in ATE the maximum deviation is just \SI{0.19}{\meter} over trajectories spanning hundreds of meters.
For object motion, the largest differences are \SI{1.29}{\degree} in rotation and \SI{0.18}{\meter} in translation.
These marginal differences, along with superior or equal performance on many sequences (\tabref{tab:object_experiments_reduced}), validate the accuracy of our proposed formulation.


\tabref{tab:object_experiments_reduced} further illustrates the impact of solver type and problem structure.
Both incremental methods show higher ATE and ME than their batch counterparts—this is a well understood outcome as iSAM2 introduces approximations (e.g., partial state updates) that can deviate the estimation from the batch-optimal solution~\cite{kaess2012isam2}.
While the performance gap between iHybrid and Hybrid can be attributed to the incremental solver, Parallel-Hybrid performs consistently worse than the methods solved jointly, regardless of solver used. This strongly affirms that solving the Dynamic SLAM problem \textit{jointly} leads to improved estimates \textit{in both camera pose and object motion}.
While prior works~\cite{bescos2021ral,Qiu2022icra_airdos} have reported similar improvements in camera pose; our results extend this by showing that the joint formulation also benefits object motion estimation.
Nevertheless, Parallel-Hybrid delivers substantial efficiency gains (\secref{sec:exp_timing}), 
therefore highlighting a practical trade-off between accuracy and computational cost.

\subsection{Incremental Solving}
\label{sec:exp_incremental_solving}
To charaterise the effect of incrementally solving each formulation we
analyse the underlying Bayes Tree's topology and assess overall computation time and scalability.
Furthermore, since the topology is determined by elimination order, we generate results with iSAM2’s \verb|relinearizeSkip| ($\lambda_{\text{rs}}$) parameter set as $1$ and $10$. This dictates when variables are relinearized which affects the portion of the Bayes Tree that is recalculated and reorded.
Setting $\lambda_{\text{rs}}=1$ enables more frequent relinearization and should result in a more optimal ordering, at the cost of greater computational time. We use $\lambda_{\text{rs}}=10$ to emphasise performance-computation trade-off, as well as the effect of sub-optimal ordering.

\subsubsection{Bayes Tree Analysis}
\label{sec:bayes_tree_analysis}
\begin{figure}[t]
    \centering
    \includegraphics[trim={0.0cm 0.0cm 0.0cm 0cm},clip,width=\linewidth]{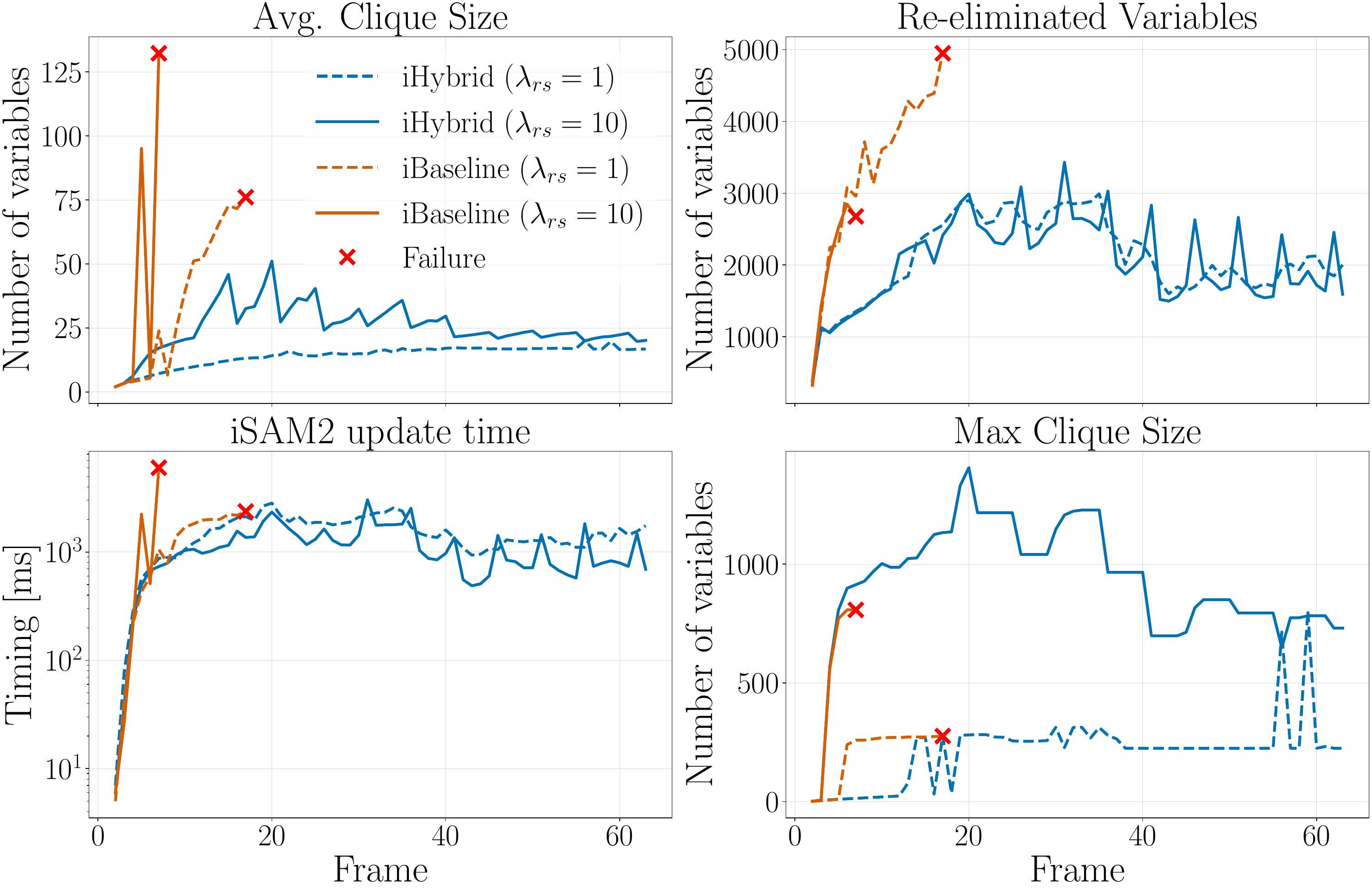}
    \cprotect\caption{\small{Bayes Tree evaluation results on a portion of KITTI $20$. 
   To demonstrate the effect of variable re-ordering and the natural scalability of the Hybrid compared to the Baseline we set $\lambda_{\text{rs}}$ to $10$ and $1$; this parameter limits relinearisation to every $n$ steps, thereby delaying variable reordering and Bayes Tree restructuring.}}
    \label{fig:kitti_0020_mem_wc_hybrid}
 \vspace{-6mm}
\end{figure}

\figref{fig:kitti_0020_mem_wc_hybrid} reports the average and maximum clique size, number of re-eliminated variables and update time of the iHybrid and iBaseline (incremental Baseline) methods.
Setting $\lambda_{\text{rs}}=1$ greatly reduces overall clique size which improves the efficiency of both methods.
However, with either $\lambda_{\text{rs}}$, iBaseline produces very large cliques and performing variable elimination eventually exceeds system memory, resulting in program failure (marked in red).
This shows the Baseline's problem structure is inherently dense and ill-suited for incremental methods.
Prior to failure, we observe a steady increase in the number of re-eliminated variables per frame - an indication that increasingly large portions of the Bayes Tree are being recomputed, which undermines the goal of incremental inference.
By comparison, Hybrid produces a Bayes Tree with consistently smaller cliques and less variables requiring re-elimination per frame, demonstrating our design is well suited for incremental solving.

\begin{figure}[t]
    \centering
    \includegraphics[trim={0.0cm 0.0cm 0.0cm 0cm},clip,width=\linewidth]{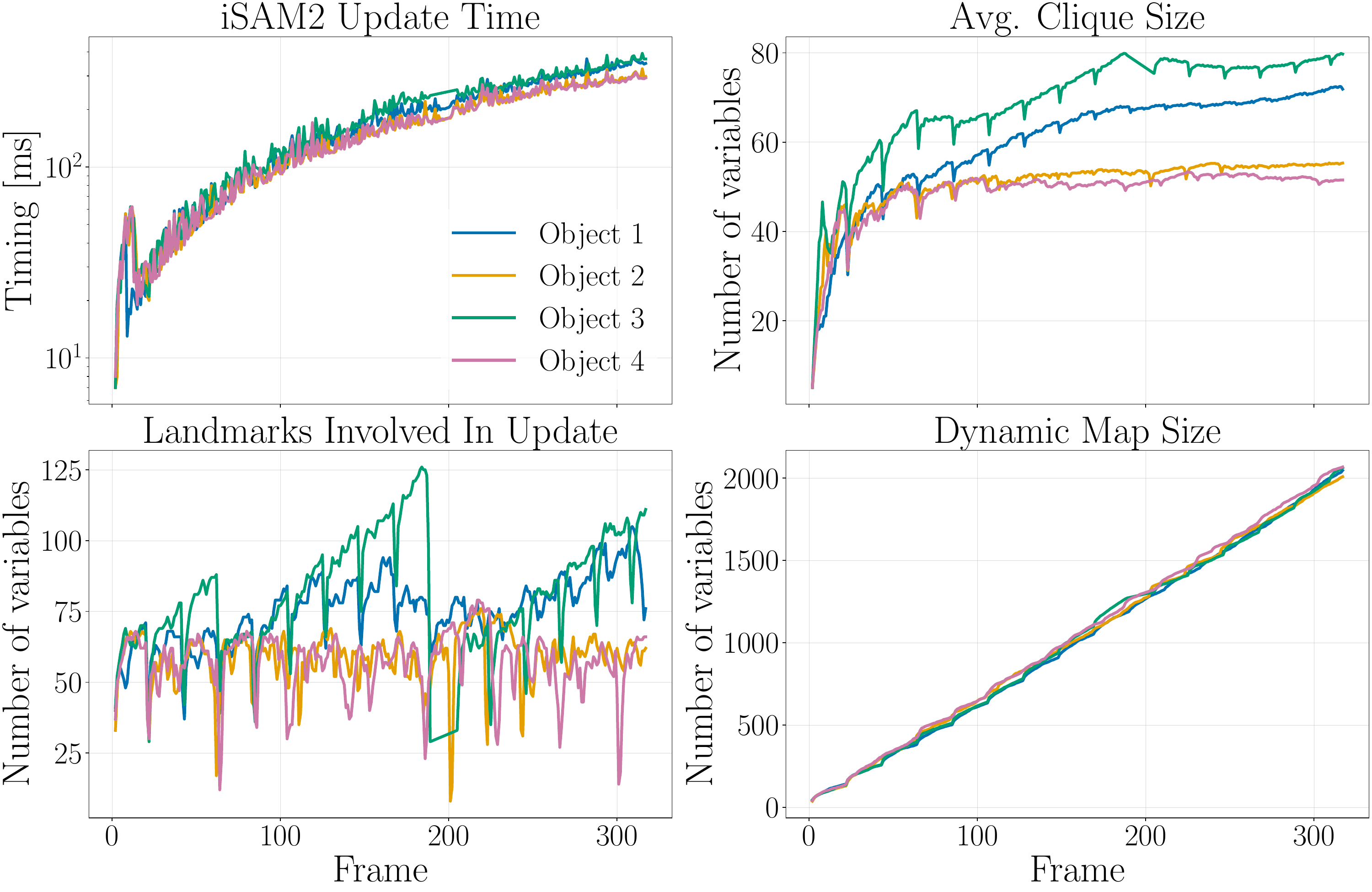}
    \cprotect\caption{\small{Evaluation results of Parallel-Hybrid on \verb|omd-s4u|: 
showing accumulated iSAM2 update time per object, average clique size, number of landmark variables involved in the update and total number of (landmark) variables in the per-object state estimate.}}
    \label{fig:omd_evaluation}
 \vspace{-4mm}
\end{figure}

\figref{fig:omd_evaluation} additionally reports the Parallel-Hybrid results on \verb|omd-s4u|, a unique sequence where all objects remain visible throughout,
thus increasing overall connectivity.
Parallel-Hybrid results in a relatively constant average clique size and a bounded update time.
We observe an interesting trend in the variable number and average clique size of objects $1$ and $3$ compared to $2$ and $4$. This corresponds with their physical trajectories: objects $1$ and $3$ exhibit less rotation relative to the camera, resulting in longer tracklets as the same faces stay visible longer which leads to the increase in clique size and spikes in affected landmarks.
This offers a key insight that in Dynamic SLAM the problem structure is impacted not only by the ego-motion, but also by objects' motion relative to the camera.
Furthermore, we note that the continuous observation of the same set of (dynamic) landmarks is similar to the special case in static SLAM where a robot remains in a restricted environment.
In that case an optimal solution can be given by eliminating all the landmark variables last via the Schur complement~\cite{kaess2008isam}. 
Given that this particular scenario is likely common in Dynamic SLAM (imagine following a car along a highway), our future work will explore using similar techniques to further improve efficiency.

\subsubsection{Timing}
\label{sec:exp_timing}

\begin{table}[t]
\footnotesize
\centering
\setlength{\tabcolsep}{3pt}
\caption{\small{iSAM2 update time for all incremental systems. Average per-frame time is reported. $\times$ indicates failure. For VIODE, Mid and High labels indicate sequence difficulty and are associated with the number of dynamic objects. We report timing results for $\lambda_{\text{rs}}$ set as $10$ and $1$.}}
\label{tab:timing_comparisons}
\begin{tabular}{ccccc}
\toprule 

\multirow{3}{*}{} & \multirow{3}{*}{} & \multicolumn{3}{c}{Timing (ms)} \\
\cmidrule{3-5}
& & Parallel-Hybrid & iHybrid & iBaseline \\
Dataset & Seq. & & \multicolumn{2}{c}{\makecell{$\lambda_{\text{rs}}=10/1$}}\\

\midrule

\multirow{9}{*}{KITTI} 
& $00$ & $\mathbf{247}$ & $\underline{601}$/$4771$ & $\times$/$2632$ \\
& $01$ & $\mathbf{232}$ & $\underline{611}$/$1683$ & $\times$/$2340$ \\
& $02$ & $\mathbf{250}$ & $\underline{648}$/$2146$ & $\times$/$\times$ \\
& $03$ & $\mathbf{333}$ & $\underline{665}$/$2202$ & $\times$/$1502$ \\
& $04$ & $\mathbf{133}$ & $\underline{733}$/$2287$ & $\times$/$3252$ \\
& $05$ & $\mathbf{332}$ & $\underline{550}$/$2931$ & $\times$/$3130$ \\
& $06$ & $\mathbf{420}$ & $1013$/$1527$ & $\times$/$\times$ \\
& $18$ & $\mathbf{397}$ &  $\times$/$8359$  &  $\times$/$7166$ \\
& $20$ & $\mathbf{1252}$ &  $\times$/$\times$  &  $\times$/$\times$  \\

\midrule

\multirow{1}{*}{OMD}
& S4U & $\mathbf{494}$ & $\times$/$\times$ & $\times$/$\times$ \\

\midrule

\multirow{4}{*}{\makecell{Outdoor\\Cluster}}
& L1 & $\mathbf{453}$ & $\underline{1268}$/$2105$ & $\times$/$3100$ \\
& L2 & $\mathbf{486}$ & $\underline{930}$/$1823$ & $1119$/$2158$ \\
& S1 & $\mathbf{124}$ & $\underline{322}$/$862$ & $330$/$868$ \\
& S2 & $\mathbf{78}$  & $\underline{244}$/$589$ & $361$/$518$ \\

\midrule

\multirow{6}{*}{\makecell{VIODE\\ {\scriptsize CD=City Day}\\{\scriptsize CN=City Night}\\{\scriptsize PL=Parking Lot}\\{\scriptsize M=Mid}\\{\scriptsize H=High}}} 
& CD\_M & $\mathbf{1332}$  & $\times$/$15822$ & $\times$/$\times$ \\
& CD\_H & $\mathbf{881}$  & $2536$/$16262$ & $\times$/$\times$  \\
& CN\_M & $\mathbf{1164}$  & $\times$/$\times$  & $\times$/$\times$  \\
& CN\_H & $\mathbf{1046}$  & $2881$/$9325$ & $\times$/$\times$ \\
& PL\_M & $\mathbf{791}$  & $\times$/$3648$ & $\times$/$\times$ \\
& PL\_H & $\mathbf{838}$  & $1947$/$4085$ & $\times$/$\times$ \\

\midrule

\multirow{4}{*}{\makecell{TartanAir\\Shibuya}}
&  IV & $\mathbf{426}$  & $\times$/$3286$ & $\times$/$\times$ \\
& V & $\mathbf{276}$  & $\underline{760}$/$1673$ & $\times$/$1031$ \\
& VI & $\mathbf{149}$  & $\underline{360}$/$663$ & $\times$/$835$ \\
& VII & $\mathbf{75}$  & $\underline{142}$/$476$ & $146$/$342$ \\

\bottomrule
\end{tabular}
\vspace{-4mm}
\end{table}

\begin{figure}[t]
    \centering
    \includegraphics[trim={0.0cm 0.0cm 0.0cm 0cm},clip,width=\linewidth]{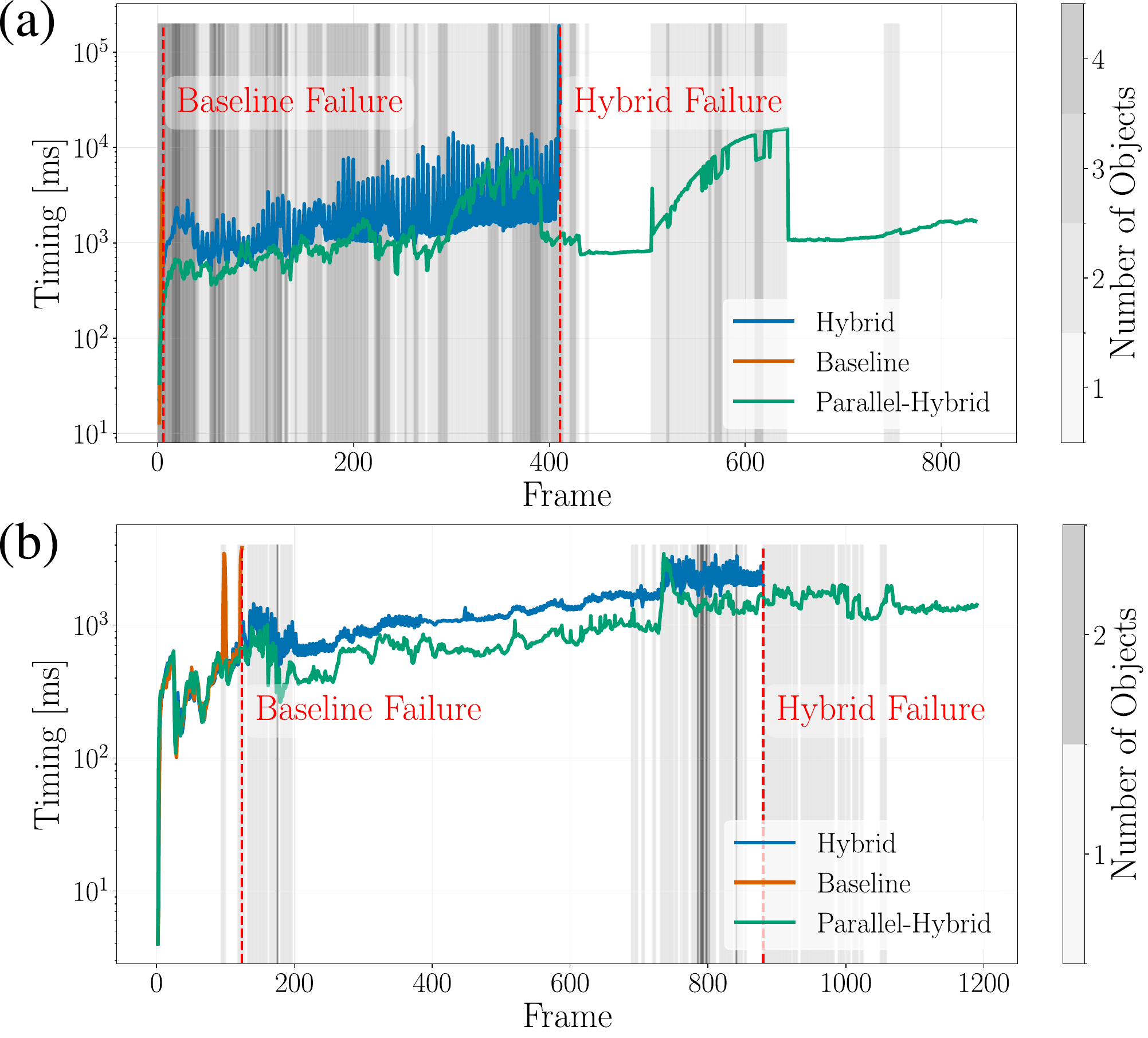}
    \cprotect\caption{\small{Per frame iSAM2 update time on selected sequences: \textbf{(a)} KITTI $20$ and \textbf{(b)} VIODE Parking Lot (Mid). The failure points are highlighted (red). Grey lines indicate the number of objects tracked at each frame.}}
    \label{fig:per_frame_timing}
 \vspace{-6mm}
\end{figure}

\tabref{tab:timing_comparisons} reports the average per-frame computation time of the iSAM2 update step.
With $\lambda_{\text{rs}}=10$ iBaseline fails on almost all sequences excluding Outdoor Cluster, which contains fewer instances of multiple objects per frame, and TartanAir Shibuya VII, where the tracks on each object are noticeably short. 
By comparison, iHybrid completes $4$ times as many sequences and is more efficient, highlight the benefit of its design.
While it still fails on long sequences with many objects, i.e. KITTI $18$ and $20$, its performance is significantly better than either iBaseline, which fails almost immediately upon observing multiple objects, as shown in~\figref{fig:per_frame_timing}.
This re-iterates the trend established in~\figref{fig:kitti_0020_mem_wc_hybrid}, where the Baseline fails due to the formation of large cliques.
While setting $\lambda_{\text{rs}}=1$ improves scalability, as both iHybrid and iBaseline methods complete more sequences, the average computation time increases by at least $200\%$ and neither method completes all sequences.

While the Hybrid structure results in significant improvements over the Baseline, the Parallel-Hybrid method is $2\times$ more efficient compared to iHybrid, $5\times$ faster than iBaseline, completes all sequences and achieves an optimisation frequency of between \SIrange{1}{5}{\hertz}.
This highlights that the object-camera connections are the key factors limiting computational efficiency. 
Overall, our results demonstrate that the Hybrid formulation is essential 
to achieve small cliques,
while the Parallel-Hybrid architecture is necessary for long-term and efficient operation.

\subsection{Real World Sequences}
To demonstrate real-world applicability, we deployed our pipeline on indoor sequences featuring multiple non-rigid (human) dynamic objects, captured using an Intel RealSense D415. 
With minimal tuning, e.g. setting a max of $100$ features per object, our method achieved online performance, with iSAM2 updates under \SI{150}{\ms}. 
As shown in \figref{fig:ecmr_frontpage}, our approach successfully estimates camera and object trajectories even with non-rigid motion, and scales to crowded scenes with up to $9$ dynamic objects, as shown in \textit{Seq1}.



\section{Future Work}
The Parallel-Hybrid's decoupled structure prevents informative dynamic observations flowing back from the DOFG's to the SFG.
Future work will enable bi-directional information exchange transfer between the dynamic and static components which is relevant when object kinematics are well-known.
Furthermore, our results indicate that solving jointly is beneficial to overall estimation accuracy. Therefore, we will continue to explore incremental methods and problem formulations that facilitate online joint estimation.




\section{Conclusion} 

We propose a novel approach to the Dynamic SLAM problem that enables online estimation by significantly enhancing problem sparsity to facilitate incremental inference.
 Our results provide insights into the structure of the Dynamic SLAM problem and reveals that decoupling the dynamic object estimation from the static scene can degrade accuracy. 
 Our proposed Hybrid formulation combines world and object-centric representations which results in a problem structure well-suited for incremental inference. 
 The Parallel-Hybrid architecture extends this formulation by enabling parallel estimation and massively improves the overall efficiency and scalability of online performance. 

\bibliographystyle{IEEEtran}
\bibliography{./IEEEabrv, ./bibliography}

\end{document}